\documentclass[10pt,conference]{IEEEtran}

\usepackage{amsmath, amssymb, amsfonts, graphicx, nicefrac, scalerel, tikz, url, cite}
\usepackage[hidelinks]{hyperref}
\usepackage[T1]{fontenc}
\usepackage{balance}

\usetikzlibrary{svg.path}
\definecolor{orcidlogocol}{HTML}{A6CE39}
\tikzset{orcidlogo/.pic={\fill[orcidlogocol]svg{M256,128c0,70.7-57.3,128-128,128C57.3,256,0,198.7,0,128C0,57.3,57.3,0,128,0C198.7,0,256,57.3,256,128z};\fill[white]svg{M86.3,186.2H70.9V79.1h15.4v48.4V186.2z}svg{M108.9,79.1h41.6c39.6,0,57,28.3,57,53.6c0,27.5-21.5,53.6-56.8,53.6h-41.8V79.1z M124.3,172.4h24.5c34.9,0,42.9-26.5,42.9-39.7c0-21.5-13.7-39.7-43.7-39.7h-23.7V172.4z}svg{M88.7,56.8c0,5.5-4.5,10.1-10.1,10.1c-5.6,0-10.1-4.6-10.1-10.1c0-5.6,4.5-10.1,10.1-10.1C84.2,46.7,88.7,51.3,88.7,56.8z};}}
\newcommand\auth[2]{#1\textsuperscript{\href{https://orcid.org/#2}{\mbox{\scalerel*{\begin{tikzpicture}[yscale=-1,transform shape]\pic{orcidlogo};\end{tikzpicture}}{|}}}}}
\newcommand\etal[1]{~\emph{et~al.}~#1}
\newcommand\tim[1]{$#1\!\times\!#1$}

\begin{document}
\title{How Unique Is a Face: An Investigative Study}
\author{
    \IEEEauthorblockN{
        \auth{Michal Balazia}{0000-0001-7153-9984}, 
        \auth{S L Happy}{0000-0003-0944-0195}, 
        \auth{Fran\c{c}ois Br\'{e}mond}{0000-0003-2988-2142}, 
        and \auth{Antitza Dantcheva}{0000-0003-0107-7029} 
    }
    \IEEEauthorblockA{
        INRIA Sophia Antipolis - M\'{e}diterran\'{e}e, 2004 Route des Lucioles, 06902 Sophia Antipolis, France
    }
    \IEEEauthorblockA{
        \texttt{\{michal.balazia, s-l.happy, francois.bremond, antitza.dantcheva\}@inria.fr}
    }
}
\maketitle


\begin{abstract}
Face recognition has been widely accepted as a means of identification in applications ranging from border control to security in the banking sector. Surprisingly, while widely accepted, we still lack the understanding of uniqueness or distinctiveness of faces as biometric modality. In this work, we study the impact of factors such as image resolution, feature representation, database size, age and gender on uniqueness denoted by the Kullback-Leibler divergence between genuine and impostor distributions. Towards understanding the impact, we present experimental results on the datasets AT\&T, LFW, IMDb-Face, as well as ND-TWINS, with the feature extraction algorithms VGGFace, VGG16, ResNet50, InceptionV3, MobileNet and DenseNet121, that reveal the quantitative impact of the named factors. While these are early results, our findings indicate the need for a better understanding of the concept of biometric uniqueness and its implication on face recognition.
\end{abstract}


\section{Introduction}
\label{intro}

\begin{figure}[ht]
\centering
\includegraphics[width=0.24\textwidth]{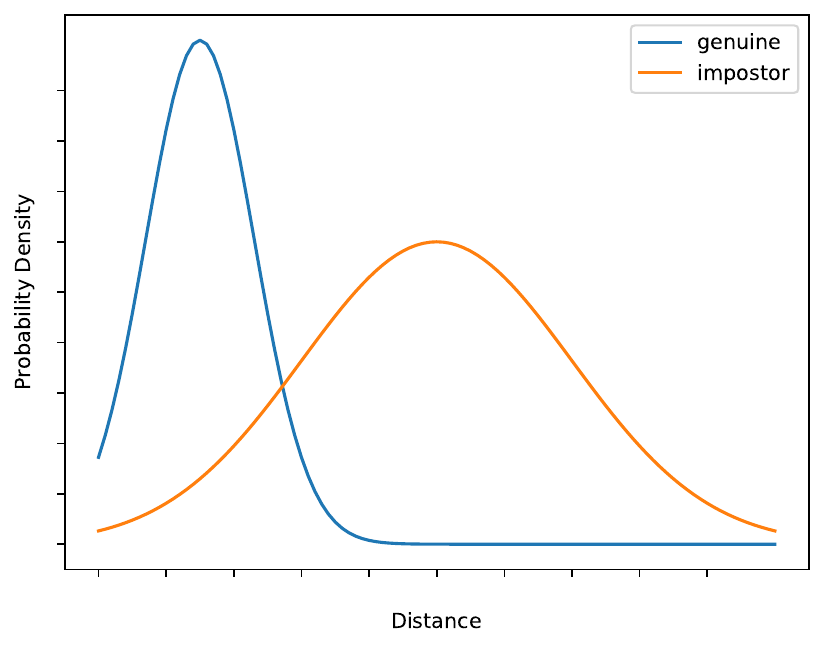}
\includegraphics[width=0.24\textwidth]{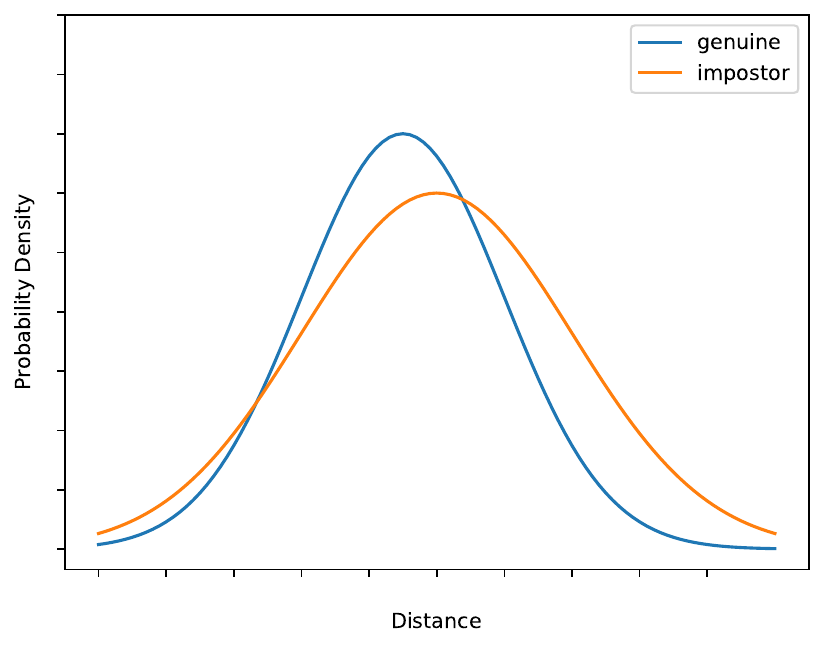}
\caption{Genuine and impostor score distributions in a setting with relatively unique subjects (left), as well as a setting with similar subjects (right).}
\label{fig-diagram}
\end{figure}

Biometrics is the science of identifying humans based on their physical, behavioral or psycho-physiological characteristics~\cite{jain2011introduction}, with one assumption being that such characteristics are unique. \textit{Uniqueness} refers to the ability of a biometric modality to distinguish between individuals, indicating how a biometric characteristic varies across the population. Consequently, an individual possesses high biometric uniqueness, if the distance distribution of their genuine (mated) comparison scores is well separated from the distance distribution of impostor (non-mated) comparison scores. Hence, an individual is less unique if their genuine and impostor distributions significantly overlap~(see Figure~\ref{fig-diagram}). We note that while a similarity match score represents a \textit{genuine score} if it is a result of matching two biometric samples of the same individual, an \textit{impostor score} refers to comparing of two biometric samples originating from different individuals~\cite{jain2007handbook}.

\begin{figure}[ht]
\centering
\includegraphics[width=0.48\textwidth]{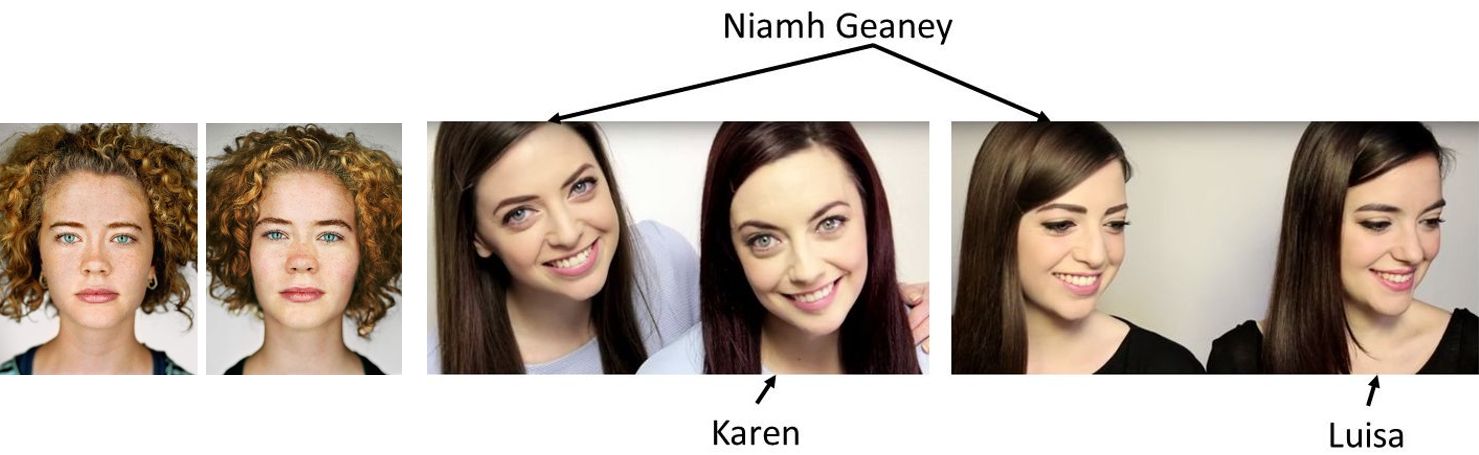}
\caption{Are faces unique? Images of identical twins (left) and doppelgangers (right). The anthropology project Twinstrangers has the goal to identify doppelgangers across the world.}
\label{fig-twins}
\end{figure}

The amount of biometric information is influenced by a set of factors including facial expression, pose, image resolution, distortion, noise or blur~\cite{daugman2015information}. Moreover, the general population includes family members, a large number of twins (3\% in the US between 2014--2018~\cite{martin2019nchs}), as well as doppelgangers\footnote{\url{https://twinstrangers.net/}}~(see Figure~\ref{fig-twins}), all of which inherently lower the overall uniqueness of faces in a dataset.

However, the knowledge of distinctiveness of face is incomplete and often relegated to anecdotal interpretation of error rates rather than a systematic exploration of the biology of the characteristics~\cite{national2009strengthening,JAIN201680,Ross19a}. Hence, we lack an estimate for the upper bound of the amount of discriminatory information contained in a face.

Such upper bound can be estimated on the level of extracted features. Generally speaking, class separability of the feature space directly corresponds to accuracy of the estimate. In the context of soft biometrics~\cite{dantcheva2015else}, distinctiveness has to do with collision, or equivalently interference, which describes the event where any two or more subjects belong in the same category of soft biometrics (e.g., female, dark hair, tall)~\cite{dantcheva2010person,dantcheva2011bag,5634534,6130409}. We note that this is related to the Birthday paradox~\cite{flajolet1992birthday} and named works answered questions such as: \emph{How large can a population become before it is likelier than not that at least two persons in the group collide biometrically?}

\begin{figure}[ht]
\centering
\includegraphics[width=0.48\textwidth]{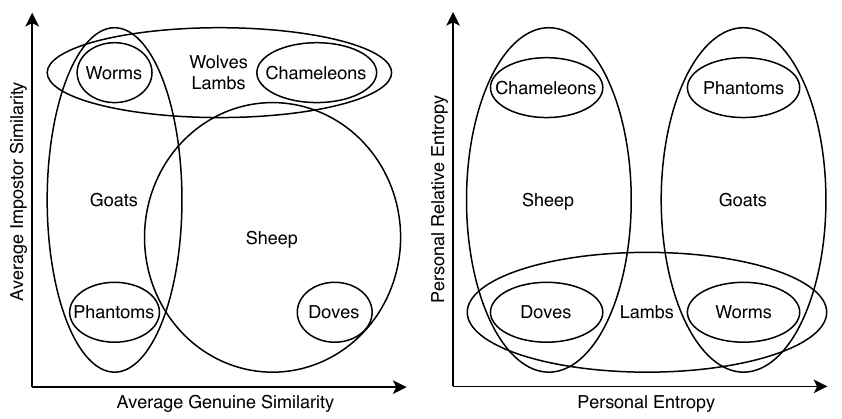}
\caption{Animal groups distinguished by Doddington's zoo according to~\cite{houmani2016hunting}. Left figure shows the distribution for match-based classification and the right one for entropy-based classification.}
\label{fig-zoo}
\end{figure}

It is known that users of a biometric system may exhibit statistically different degrees of accuracy within the system, which relates to variations in uniqueness across subjects. While some users may experience challenges in authentication, others may be particularly vulnerable to impersonation. The Doddington's zoo~\cite{doddington1998sheep,poh2006revisiting,ross2009exploiting} has been frequently used to quantify this phenomenon and specifically to classify users based on verification performance, when users are compared against themselves and against others. Associated major classes include: (a)~\emph{sheep}: users who are easy to recognize; (b)~\emph{goats}: users who are difficult to recognize; (c)~\emph{lambs}: users who are easy to imitate; and (d)~\emph{wolves}: users who can easily imitate others. Easy imitation contributes to False Acceptance Rate and difficult recognition to False Rejection Rate. This concept was extended in the biometric menagerie~\cite{houmani2016hunting}~(see Figure~\ref{fig-zoo}), with additional classes: (e)~\emph{chameleons}: users who are easy to recognize and easy to imitate; (f)~\emph{phantoms}: users who are difficult to recognize and difficult to imitate; (g)~\emph{doves}: users who are easy to recognize and difficult to imitate; and (h)~\emph{worms}: users who are difficult to recognize and easy to imitate. Hence, the distribution of a given dataset in such classes impacts the accuracy of a face recognition system on the dataset.

Motivated by the above, we here aim to provide new insight on the topic of facial uniqueness by investigating the impact of feature extractors (in particular six CNN-based feature extractors: VGGFace, VGG16, ResNet50, InceptionV3, MobileNet and DenseNet121), dataset size (see Section~\ref{datasets}), image resolution (see Section~\ref{resolution}), as well as age and gender (see Section~\ref{subsets}), on the four datasets AT\&T, LFW, IMDb and TWINS. Additionally, we identify a potential limitation of the used uniqueness score and propose an additional uniqueness score, which reflects the common occurrence of persons, who look alike (twins, family members and doppelgangers).


\section{Related Work}
\label{rel_work}

There have been numerous early attempts to quantify biometric uniqueness, distinctiveness and entropy. A key estimator of biometric information related to uniqueness was defined by Adler\etal{\cite{adler2006towards}} as \emph{the decrease in uncertainty about the identity of a person due to a set of biometric measurements}. This measure was subsequently used by Sutcu\etal{\cite{sutcu2013biometric}}, Takashi and Murakami~\cite{takahashi2010metric}, as well as Gomez-Barrero\etal{\cite{gomez2017biometrie}}. Similar uniqueness considerations were applied towards improving template protection~\cite{gomez2017biometrie} and performance of biometric systems~\cite{klare2012face}. Further, facial entropy was quantified and was found to range from $2.6$~bits~\cite{o2003comparing} to $55$~bits~\cite{adler2006towards}. Recent work of Krivokuća and Marcel~\cite{krivokuca2018fingervein} on fingervein entropy estimated discriminatory information in fingervein patterns by calculating the number of features used to differentiate between different fingers. 

While such works have used different measures of biometric information, what they have in common is the calculation of entropy, which relates to the inter-class variance that can be artificially increased by including a larger number of features. In order to be independent of the number of extracted features and thus to more accurately estimate the intrinsic information, we also need to consider intra-class variance. Both of these variances are encompassed in our proposed biometric uniqueness score based on the Kullback-Leibler divergence~\cite{kullback1951kld}, also denoted as \textit{relative entropy}, between genuine and impostor distributions. By its definition below, KL-divergence weighs resistance to both false acceptance and false rejection.

In the context of iris and iris code in particular, an iris uniqueness has been determined by Daugman~\cite{daugman2006probing,daugman2015information}. Therein $28\%$ of the bits in two IrisCodes are allowed to disagree while still accepting them as a match, resulting in a false match rate of $1$ in $92$ billion. Given the fixed length of the standardized IrisCode, we consider this rate as an exact indicator of uniqueness in irises.

For the purpose of texture-based generation of fingerprint images, Yankov\etal{\cite{yankov2020fingerprint}} used generative models to estimate upper bounds on the image entropy for systems with small sensor acquisition. They estimated the identification capacity of such systems using the mutual information between different samples from the same finger.

We proceed with describing the uniqueness measure to quantify the identification capacity of faces based on a similar concept.


\section{Uniqueness as a Divergence}
\label{uniq}

Takahashi\etal{\cite{Takahashi14}} showed that a decrease in uncertainty with respect to identity of an unknown biometric characteristic can be formulated in terms of mutual information
\begin{equation}
    \mathit{I}\!\left(X;Y\right)=H\!\left(X\right)-H\!\left(X|Y\right)
\end{equation}
with $H\!\left(X\right)$ as the marginal entropy, i.e., uncertainty of $X$, and $H\!\left(X|Y\right)$ as the conditional entropy, i.e., uncertainty of $X$ given the observation of $Y$. In addition, the authors showed that $\mathit{I}\!\left(X;Y\right)$ can be approximated by the Kullback-Leibler divergence of genuine $P_G(d)$ and impostor $P_I(d)$ probability distributions
\begin{equation}
    \mathit{I}\!\left(X;Y\right) \approx \mathrm{D}\!\left(P_G\bigl\|P_I\right) = \sum_d P_G\!\left(d\right)\mathrm{log}\frac{P_G\!\left(d\right)}{P_I\!\left(d\right)},
\end{equation}
where $d$ denote observed dissimilarities between pairs of samples constituting a genuine or impostor observations.

We here note that an average norm estimator can be used to calculate the value of $\mathrm{D}\!\left(P_G\bigl\|P_I\right)$ from the dissimilarities of samples without computing any probability models~\cite{Sutcu10}. Let $G$ and $I$ be i.i.d. sets of samples with mutual dissimilarities forming the distributions $P_G$ and $P_I$, respectively. Then, the average norm estimator of the KL-divergence is defined as
\begin{equation}
    \mathrm{D}\!\left(P_G\bigl\|P_I\right) \approx \hat{\mathrm{D}}\!\left(G,I\right) = \frac{1}{\left|G\right|} \sum_{g \in G} \mathrm{log}\frac{\delta_g\!\left(I\right)}{\delta_g\!\left(G\right)} + \mathrm{log}\frac{\left|I\right|}{\left|G\right|-1},
    \label{eq-kldiv}
\end{equation}
where
\begin{equation}
    \delta_s\!\left(S\right) = \frac{1}{\left|S\setminus\left\{s\right\}\right|} \sum_{s' \in S\setminus\left\{s\right\}} \left\lVert s-s' \right\rVert_2
\end{equation}
represents mean of the Euclidean norms between a sample $s$ and all samples from the set $S$ eventually without $s$.

Calculation of $\mathrm{D}\!\left(P_G\bigl\|P_I\right)$ can be sped up by approximation with random sampling. In order to ensure that the score is not biased towards the distribution of a certain random subset, we calculate the divergence as the average of $n$ different random subset choices. We note that random samples are a sounder representation of the underlying distribution as compared to for example nearest neighbors. Furthermore, for the divergence to mainly depend on the genuine-impostor distributions rather than the count distribution, we choose the subsets $G'\!\subseteq\!G$ and $I'\!\subseteq\!I$ to contain $r+1$ and $r$ random samples from $G$ and $I$, respectively. Equation~\ref{eq-kldiv} then transforms to
\begin{equation}
    \begin{split}
    \hat{\mathrm{D}}_{n,r}\!\left(G,I\right) & = \frac{1}{n} \sum_{k=1}^n \left(\frac{1}{r} \sum_{g \in G'} \mathrm{log}\frac{\delta_g\!\left(I'\right)}{\delta_g\!\left(G'\right)} + \mathrm{log}\frac{r}{r}\right) \\
    & = \frac{1}{nr} \sum_{k=1}^n \sum_{g \in G'} \mathrm{log}\frac{\delta_g\!\left(I'\right)}{\delta_g\!\left(G'\right)}.
    \end{split}
\end{equation}

In our experiments, we calculate KL-divergence from the above definition using all available genuine data
\begin{equation}
    r=\mathrm{min}\!\left(\left|G\right|-1,\left|I\right|\right)
\end{equation}
so as to obtain the most accurate approximation of the genuine and impostor distributions and
\begin{equation}
    n=\lceil\nicefrac{100}{r}\rceil
\end{equation}
random subset choices which we believe is sufficiently large to prevent a potential bias. For very large datasets that have too many samples of individual subjects, one can take a smaller $r$ to estimate the divergence for reasons of efficiency. Slightly abusing annotation, in the following we will refer to $\hat{\mathrm{D}}_{\lceil\nicefrac{100}{\mathrm{min}\left(\left|G\right|-1,\left|I\right|\right)}\rceil,\mathrm{min}\left(\left|G\right|-1,\left|I\right|\right)}$ as to simply $\hat{\mathrm{D}}$.

Let a dataset $S$ have $c$ subjects and let $S_p\!\subset\!S$ denote a set of samples that belong to subject $p$. The KL-divergence estimate on this dataset is defined as the average $\hat{\mathrm{D}}$ across all subjects
\begin{equation}
    \bar{\mathrm{D}}\!\left(S\right) = \frac{1}{c}\sum_{p=1}^c \hat{\mathrm{D}}\!\left(S_p,S\!\setminus\!\left\{S_p\right\}\right),
    \label{eq-div}
\end{equation}
where impostor distribution is calculated from the whole remainder of the dataset $S\!\setminus\!\left\{S_p\right\}$.

Finally, since there is a common practice of normalizing biometric scores to the interval $(0,1)$, we define the impostor-based biometric uniqueness $\mathrm{U}$ of a given dataset $S$ as the estimated and sigmoid-normalized divergence
\begin{equation}
    \mathrm{U}\!\left(S\right) = \frac{1}{1+\mathrm{e}^{-\bar{\mathrm{D}}\left(S\right)}}.
    \label{eq-uniq}
\end{equation}
Note that while $\mathrm{U}\!\left(S\right)$ close to 1 indicates that subjects are highly unique, a lower uniqueness indicates that the genuine and impostor distributions are increasingly overlapping. Also note that the uniqueness score is defined as an exponential function with base $\mathrm{e}$, which means that even a small variation in the uniqueness score has a rather significant effect.


\section{Datasets}
\label{dat}

In this section we briefly present the four used datasets: AT\&T, LFW, IMDb and TWINS. We summarize details in Table~\ref{dat-t}.

\begin{itemize}\setlength\itemsep{5pt}
\item \textbf{AT\&T Database of Faces}~(AT\&T)~\cite{samaria1994att} contains face images acquired in the laboratory conditions, with a dark homogeneous background. Subjects are captured in an upright, frontal position. There are 10 different images of each of 40 subjects. Covariates include lighting, facial expressions, open or closed eyes, as well as facial accessories such as glasses. 
\item{\textbf{Labeled Faces in the Wild}} (LFW)~\cite{huang2007lfw} was designed for studying the problem of unconstrained face recognition. The dataset contains 5,749 distinct subjects with 13,233 images of faces collected from the web and hence incorporates highly unconstrained conditions. We select 9,164 images of 1,680 subjects, with two or more images in the dataset. Faces were detected and cropped by the Viola-Jones face detector~\cite{viola2004detector}.
\item{\textbf{IMDb-Face}} (IMDb)~\cite{wang2018imdb} is a large-scale noise-controlled dataset for face recognition research, containing about 1.7M faces of 59K identities, which is a manually cleaned subset of the original 2M raw images. All images were obtained from the IMDb website. We select 1,167,509 images associated to 10,347 identities. IMDb contains gender and age annotations, which allow us to perform additional experiments on subsets pertained to gender and age. See statistics in Table~\ref{dat-tx}.
\item{\textbf{ND-TWINS-2009-2010}} (TWINS)~\cite{phillips2011twins} dataset was collected by the University of Notre Dame in the years 2009-2010 and comprises 24,050 color photographs of 435 attendee faces captured at the Twins Days Festival under natural light in indoor and outdoor configurations. Facial yaw varies from $-90$ to $+90$ degrees in steps of $45$ degrees. We use a set of cropped 23,762 images, where we detect faces with the MTCNN detector~\cite{zhang2016mtcnn}.
\end{itemize}

\begin{table}[ht]
\caption{Sizes of datasets with mean and standard deviation statistics, gender and ethnicity distributions.}
\label{dat-t}
\centering
\setlength{\tabcolsep}{3pt}
\begin{tabular}{|r|rrrr|}
\hline
& AT\&T & LFW & IMDb & TWINS \\
\hline
total number of subjects & 40 & 1,680 & 10,347 & 435 \\
total number of samples & 400 & 9,164 & 1,167,509 & 23,762 \\
mean/st.d. samples per subject & 10.0/0.0 & 5.5/16.3 & 112.8/70.5 & 54.6/36.1 \\
\% female/male & 10/90 & 23/77 & 40/60 & 75/25 \\
\% African/Asian/Caucasian & 2/0/98 & 9/8/83 & 12/23/65 & 12/1/84 \\
\hline
\end{tabular}
\end{table}

\begin{table}[ht]
\caption{Numbers of subjects in IMDb by age and gender.}
\label{dat-tx}
\centering
\begin{tabular}{|rrrrrrr|}
\hline
full & male & female & 0-9 & 10-19 & 20-29 & 30-39 \\
\hline
10,347 & 6,215 & 4,132 & 159 & 856 & 2558 & 3031 \\
\hline\hline
40-49 & 50-59 & 60-69 & 70-79 & 80-89 & 90-99 & 100-109 \\
\hline
1876 & 1024 & 522 & 227 & 77 & 14 & 2 \\
\hline
\end{tabular}
\end{table}


\section{Algorithms}
\label{alg}

We select six state-of-the-art convolutional neural networks (CNNs), which have excelled in a number of face recognition and classification tasks that we proceed to describe.
\begin{itemize}
\item{\textbf{VGGFace}}~\cite{parkhi2015vggface} is a 16-layer CNN, trained on the VGGFace dataset comprising over 2M celebrity images. Given a \tim{224} input image, the network extracts 4,096 image features from the output of the 6-th fully connected layer.
\item{\textbf{VGG16}}~\cite{simonyan2016vgg} constitutes the same 16-layer CNN, however trained on over 1M images from the ImageNet database~\cite{russakovsky2014imagenet}.
We use 4,096 features, provided as output by the second fully connected layer.
\item{\textbf{ResNet50}}~\cite{he2016resnet} represents a 50-layer CNN, also trained on ImageNet. As a result of having a large amount of layers, the network has learned rich feature representations for a wide range of images. Similar as the above networks, ResNet50 accepts as input an image of size \tim{224}. The 2,048 related ResNet50 features are provided by the last convolutional layer.
\item{\textbf{InceptionV3}}~\cite{szegedy2016inception} is a 48-layer CNN trained on ImageNet. We obtain 2,048 features from the last fully connected layer.
\item{\textbf{MobileNet}}~\cite{howard2017mobilenet} is a 56-layer CNN trained on ImageNet, which has been optimized for mobile devices. MobileNet extracts a 1,000-dimensional feature vector.
\item{\textbf{DenseNet121}}~\cite{huang2017densenet} is a 121-layer CNN trained on ImageNet. The layer structure involves more narrow layers as opposed to ResNet50. The total number of layers is determined by 5 plus two blocks of 58 layers, the last of which results in a 1,024-dimensional feature vector.
\end{itemize}


\section{Experiments}
\label{exp}

We conduct a set of experiments on the above enlisted datasets with the above named algorithms. We report in each experiment the uniqueness score $\mathrm{U}$, as denoted in Equation~\ref{eq-uniq}. We note that small variations in $\mathrm{U}$ can indicate large differences in uniqueness due to the exponential function in the denominator in Equation~\ref{eq-div}. We proceed to investigate $\mathrm{U}$ with respect to following factors.

\subsection{Datasets}
\label{datasets}

In the first experiment, we study the uniqueness score across datasets of different size and setting. Related scores are reported in Table~\ref{exp-t-full}. We note that dataset size is pertinent, as in large datasets it is more likely to encounter similar faces, and the related probability of collision is higher as opposed to small datasets. As expected, AT\&T encompasses higher uniqueness scores, due to its small size, as well as constrained acquisition conditions. With respect to features, we see that VGGFace features are systematically outperforming the other networks w.r.t. uniqueness and hence distinctiveness. We believe it is because VGGFace was trained on faces as opposed to other models trained on ImageNet.

\begin{table}[ht]
\caption{Uniqueness evaluated on full AT\&T, LFW, IMDb and TWINS, with resolution \tim{224}.}
\label{exp-t-full}
\centering
\begin{tabular}{|r|llll|}
\hline
& AT\&T & LFW & IMDb & TWINS \\
\hline
VGGFace & 0.6710 & 0.5637 & 0.5420 & 0.5591 \\
VGG16 & 0.6417 & 0.5364 & 0.5381 & 0.5339 \\
ResNet50 & 0.6650 & 0.5364 & 0.5330 & 0.5574 \\
InceptionV3 & 0.5915 & 0.5293 & 0.5272 & 0.5242 \\
MobileNet & 0.6204 & 0.5353 & 0.5299 & 0.5301 \\
DenseNet121 & 0.6338 & 0.5302 & 0.5309 & 0.5268 \\
\hline
\end{tabular}
\end{table}

\subsection{Image Resolution}
\label{resolution}

We proceed to investigate five image resolutions, namely \tim{224}, \tim{112}, \tim{64}, \tim{48} and \tim{36}, of the cropped face images of our datasets, all of them in RGB. Given that some of the CNNs take an input of size \tim{224}, we apply a lossy data conversion by first downscaling the image to a given resolution and subsequently upscaling it to \tim{224}, in order to fit the standardized input size. We observe that resolution has a surprisingly low effect on uniqueness, as shown in Table~\ref{exp-t-res}. 

For this experiment we provide an additional measure related to uniqueness, namely the image entropy calculated by
\begin{equation}
    H = -\sum_{c=1}^C p_c \cdot log_2\left(p_c\right),
\end{equation}
where $C$ is the color depth, $p_c$ is probability of color $c$, that is, number of pixels of color $c$ by total number of pixels. We see that while $H$ decreases proportionally with resolution, $\mathrm{U}\!\left(S\right)$ is not substantially affected.

\begin{table}[ht]
\caption{Uniqueness (and mean image entropy) evaluated on full AT\&T, LFW, IMDb and TWINS, with VGGFace feature extraction algorithm.}
\label{exp-t-res}
\centering
\setlength{\tabcolsep}{4pt}
\begin{tabular}{|r|llll|}
\hline
& AT\&T & LFW & IMDb & TWINS \\
\hline
\tim{224} & 0.6710 (68.8) & 0.5637 (87.4) & 0.5420 (43.9) & 0.5591 (27.8) \\
\tim{112} & 0.6725 (64.0) & 0.5636 (70.1) & 0.5418 (38.1) & 0.5590 (25.3) \\
\tim{64} & 0.6620 (60.4) & 0.5635 (59.3) & 0.5405 (32.1) & 0.5581 (23.3) \\
\tim{48} & 0.6485 (55.5) & 0.5637 (34.4) & 0.5385 (27.9) & 0.5566 (21.9) \\
\tim{36} & 0.6241 (51.2) & 0.5603 (19.0) & 0.5345 (20.4) & 0.5522 (19.7) \\
\hline
\end{tabular}
\end{table}

\begin{figure*}[ht]
\centering
\includegraphics[width=0.99\textwidth]{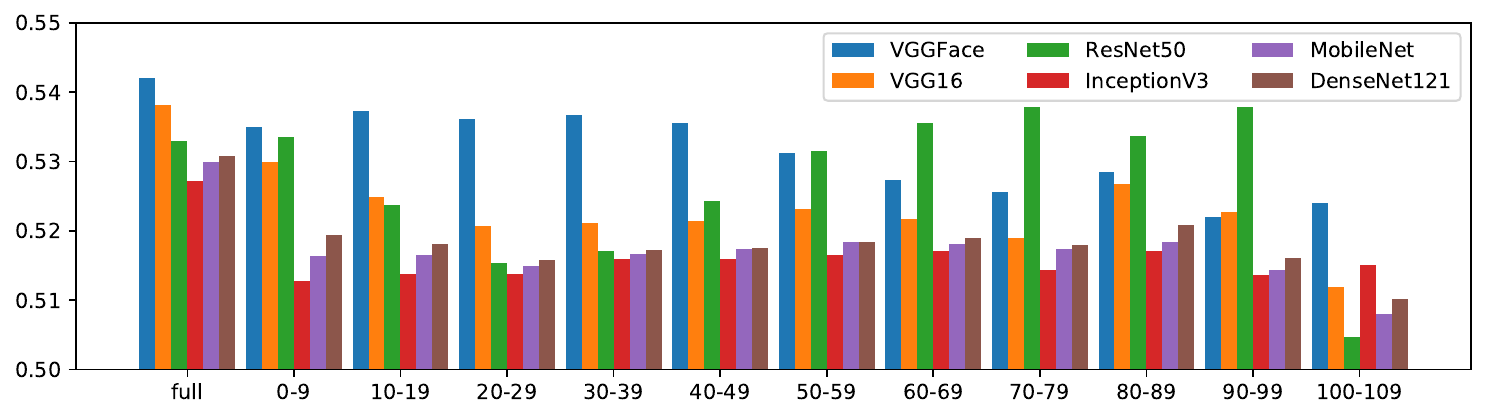}
\caption{Uniqueness evaluated on full IMDb compared to its split into 10-year blocks.}
\label{exp-f-age}
\end{figure*}

\subsection{Subsets}
\label{subsets}

We investigate the effect of \textit{gender} and \textit{age} on the uniqueness score in the dataset IMDb, since related annotations are provided. Results for the gender split are depicted in Table~\ref{exp-t-gender}. Intuitively, the full dataset reflects a higher uniqueness than the gender subsets, which are likely to include more similar faces of only females and only males.

A similar claim can be made w.r.t. age split. Results in Figure~\ref{exp-f-age} show a decrease in uniqueness in all 10-year age-subsets compared to the full IMDb dataset. Moreover, we observe a slight trend of lower uniqueness scores associated with all algorithms for infants and seniors, and of higher uniqueness for middle-aged people. This can be related to the other-age-effect observed in psychology \cite{kuefner2008all}.  

\begin{table}[ht]
\caption{Uniqueness evaluated on full IMDb with resolution \tim{224} compared to its splits into genders.}
\label{exp-t-gender}
\centering
\begin{tabular}{|r|lll|}
\hline
& full & female & male \\
\hline
VGGFace & 0.542 & 0.535 & 0.532 \\
VGG16 & 0.538 & 0.520 & 0.521 \\
ResNet50 & 0.533 & 0.519 & 0.525 \\
InceptionV3 & 0.527 & 0.511 & 0.514 \\
MobileNet & 0.530 & 0.514 & 0.516 \\
DenseNet121 & 0.531 & 0.515 & 0.517 \\
\hline
\end{tabular}
\end{table}


\section{Limitation of \texorpdfstring{$\bar{\mathrm{U}}$}{U} and Introduction of \texorpdfstring{$\bar{\mathrm{U}}^\mathrm{MIN}$}{Umin}}
\label{uniq_min}

We recall the definition of the uniqueness score in Equation~\ref{eq-div}, where impostors are randomly drawn from the whole remainder of the dataset. Under this definition, given a dataset of N subjects, replacing half of the subjects with twins of the already included subjects, would lower the divergence by a factor of $\nicefrac{1}{N}$. Hence the divergence of a dataset of 100 twins will be about 99\% of the divergence of non-twin dataset, which is negligible and therefore represents a limitation of the used divergence estimate $\bar{\mathrm{D}}$. For an application that requires a significant impact of occurrence of twins on the uniqueness score, we propose a revised divergence estimate $\bar{\mathrm{D}}^\mathrm{MIN}$, which places emphasis on the most similar impostor instead of all remaining people. This amendment moves the impostor distribution towards genuine mildly for general population datasets and significantly for twin datasets.

Let a dataset $S$ have $c$ subjects and $S_p\!\subset\!S$ denote a set of samples that belong to the subject $p$. As an alternative, the minimum KL-divergence estimate on this dataset is defined as the minimum $\hat{\mathrm{D}}$ across all pairs of subjects
\begin{equation}
    \bar{\mathrm{D}}^\mathrm{MIN}\!\left(S\right) = \frac{1}{c}\sum_{p=1}^c \mathrm{min}_{\genfrac{}{}{0pt}{2}{q = 1}{q \neq p}}^c \hat{\mathrm{D}}\!\left(S_p,S_q\right),
    \label{eq-div-min}
\end{equation}
where impostor distribution is calculated from the samples $S_q$ of only the subject $q$ with the closest distribution to $S_p$. The corresponding alternative uniqueness scores $\mathrm{U}^\mathrm{MIN}$, calculated using $\bar{\mathrm{D}}^\mathrm{MIN}$ as
\begin{equation}
    \mathrm{U}^\mathrm{MIN}\!\left(S\right) = \frac{1}{1+\mathrm{e}^{-\bar{\mathrm{D}}^\mathrm{MIN}\left(S\right)}},
    \label{eq-uniq-min}
\end{equation}
are reported in Table~\ref{exp-t-dmin}. We observe that while the AT\&T dataset based on $\bar{\mathrm{D}}^\mathrm{MIN}$ has a slightly lower $\mathrm{U}^\mathrm{MIN}\!\left(\mathrm{AT\&T}\right)$ score than $\mathrm{U}\!\left(\mathrm{AT\&T}\right)$, the TWINS dataset drops $\bar{\mathrm{D}}^\mathrm{MIN}\!\left(\mathrm{TWINS}\right)\approx0$ and so $\mathrm{U}^\mathrm{MIN}\!\left(\mathrm{TWINS}\right)\approx0.5$, as desired.

\begin{table}[ht]
\caption{Uniqueness evaluated on \tim{224} resolution with original and minimum divergence estimates.}
\label{exp-t-dmin}
\centering
\begin{tabular}{|r|ll|ll|}
\hline
& \multicolumn{2}{|c|}{$\mathrm{U}$} & \multicolumn{2}{|c|}{$\mathrm{U}^\mathrm{MIN}$} \\
\hline
& AT\&T & TWINS & AT\&T & TWINS \\
\hline
VGGFace & 0.671 & 0.559 & 0.632 & 0.507 \\
VGG16 & 0.642 & 0.534 & 0.582 & 0.491 \\
ResNet50 & 0.665 & 0.557 & 0.542 & 0.449 \\
InceptionV3 & 0.592 & 0.524 & 0.542 & 0.485 \\
MobileNet & 0.620 & 0.530 & 0.570 & 0.498 \\
DenseNet121 & 0.634 & 0.527 & 0.577 & 0.494 \\
\hline
\end{tabular}
\end{table}


\section{Conclusions}
\label{conclusions}

In this work we presented preliminary results on the impact of factors such as image resolution, gender, age, datasets, as well as feature extraction algorithms on facial uniqueness. This is the first work that systematically studies such factors. We provided clear experimental evidence of decrease in the uniqueness score, in the case that (a)~image resolution decreases, (b)~a single gender is observed, (c)~a smaller age group is observed, (d)~a larger dataset is used, as well as (e)~different feature extractors are used. We illustrated that while feature representation and dataset size significantly affect the uniqueness score, image resolution has a negligible impact. Further, we proposed an alternative uniqueness estimate, which reflects on the presence of twins. Future work will involve establishing a more detailed experimental protocol that among others will aim at quantifying the impact of facial symmetry on uniqueness. Further, common notions of facial entropy, distinctiveness, diversity, complexity, averageness and attractiveness, and their associated relations are to be explored.


\section*{Acknowledgment}

This work was supported by the French National Research Agency (ANR) under grants ENVISION ANR-17-CE39-0002 and IDEX UCA\textsuperscript{JEDI} ANR-15-IDEX-01.


\bibliographystyle{plain}
\balance
\bibliography{ref}
\end{document}